\documentclass[journal]{IEEEtran}

\ifCLASSINFOpdf
\else
   \usepackage[dvips]{graphicx}
\fi
\usepackage{url}

\usepackage{graphicx}
\usepackage{enumitem}
\usepackage{amsmath}
\usepackage{amssymb}
\usepackage{booktabs}
\usepackage{makecell} 
\usepackage{colortbl} 
\usepackage{multirow}
\usepackage{xcolor}

\begin{document}

\title{Click on Mask: \\A Labor-efficient Annotation Framework with Level Set for Infrared Small Target Detection}

\author{Haoqing Li, Jinfu Yang, Yifei Xu, Runshi Wang
\thanks{This work was supported in part by the National Natural Science Foundation of China under Grant no.61973009. (Corresponding author: Jinfu Yang.) 
	
	Haoqing Li, Runshi Wang, and Yifei Xu are with the Faculty of Information, Beijing University of Technology, Beijing, 100124, P. R. China. (e-mails: lihaoqing@emails.bjut.edu.cn; wangrunshi@emails.bjut.edu.cn; xuyifei@emails.bjut.edu.cn). 
	
	Jinfu Yang is with the Faculty of Information Technology, Beijing University of Technology, Beijing, 100124, China, also with the Beijing Key Laboratory of Computational Intelligence and Intelligent System, Beijing University of Technology, Beijing, 100124, P. R. China (e-mail: jfyang@bjut.edu.cn). }}

\markboth{Journal of \LaTeX\ Class Files, Vol. 14, No. 8, August 2015}
{Shell \MakeLowercase{\textit{et al.}}: Bare Demo of IEEEtran.cls for IEEE Journals}
\maketitle

\begin{abstract}
Infrared Small Target Detection is a challenging task to separate small targets from infrared clutter background. Recently, deep learning paradigms have achieved promising results. However, these data-driven methods need plenty of manual annotation. Due to the small size of infrared targets, manual annotation consumes more resources and restricts the development of this field. This letter proposed a labor-efficient and cursory annotation framework with level set, which obtains a high-quality pseudo mask with only one cursory click. A variational level set formulation with an expectation difference energy functional is designed, in which the zero level contour is intrinsically maintained during the level set evolution. It solves the issue that zero level contour disappearing due to small target size and excessive regularization. Experiments on the NUAA-SIRST and IRSTD-1k datasets reveal that our approach achieves superior performance. Code is available at https://github.com/Li-Haoqing/COM.
\end{abstract}


\IEEEpeerreviewmaketitle

\section{Introduction}

\IEEEPARstart{I}{nfrared} Small Target Detection (IRSTD), a task to separate small targets from infrared clutter background, provides many potential applications in remote sensing, public security, and other fields \cite{1}, \cite{2}, \cite{3}, \cite{4}. The small area ($<$ 15 × 15 pixels) and low signal-to-clutter ratio make the IRSTD an extremely challenging task. Conventional non-learning approaches, for instance, filtering-based \cite{5}, \cite{6}, local contrast-based \cite{7}, \cite{8}, and low rank-based \cite{9}, \cite{10}, \cite{11}, \cite{12}, \cite{13} approaches are hindered by numerous hyper-parameters, high false alarm rate and missing detection. In addition, deep learning approaches have achieved better pixel-level accuracy, lower false alarm and miss detection. Dai \textit{et al.} \cite{14}, \cite{15} proposed an asymmetric contextual modulation module to exchange high-level semantics and subtle low-level details. Li \textit{et al.} \cite{16} proposed a dense nested attention network to incorporate and exploit contextual information. Zhang \textit{et al.} \cite{17} emphasized that shape matters and incorporated shape reconstruction into IRSTD.

At present, the mainstream supervised learning methods require a large amount of data for model training, but pixel-level annotation is extremely labor-intensive and time-consuming. Further, because of the small size and inferior separability, each pixel has a greater impact on the results. Therefore, the mask of IRSTD data needs pixel-level annotating meticulously in the enlarged image region, which consumes more resources.

\begin{figure}[!t]
	\centering
	\includegraphics[width=\columnwidth]{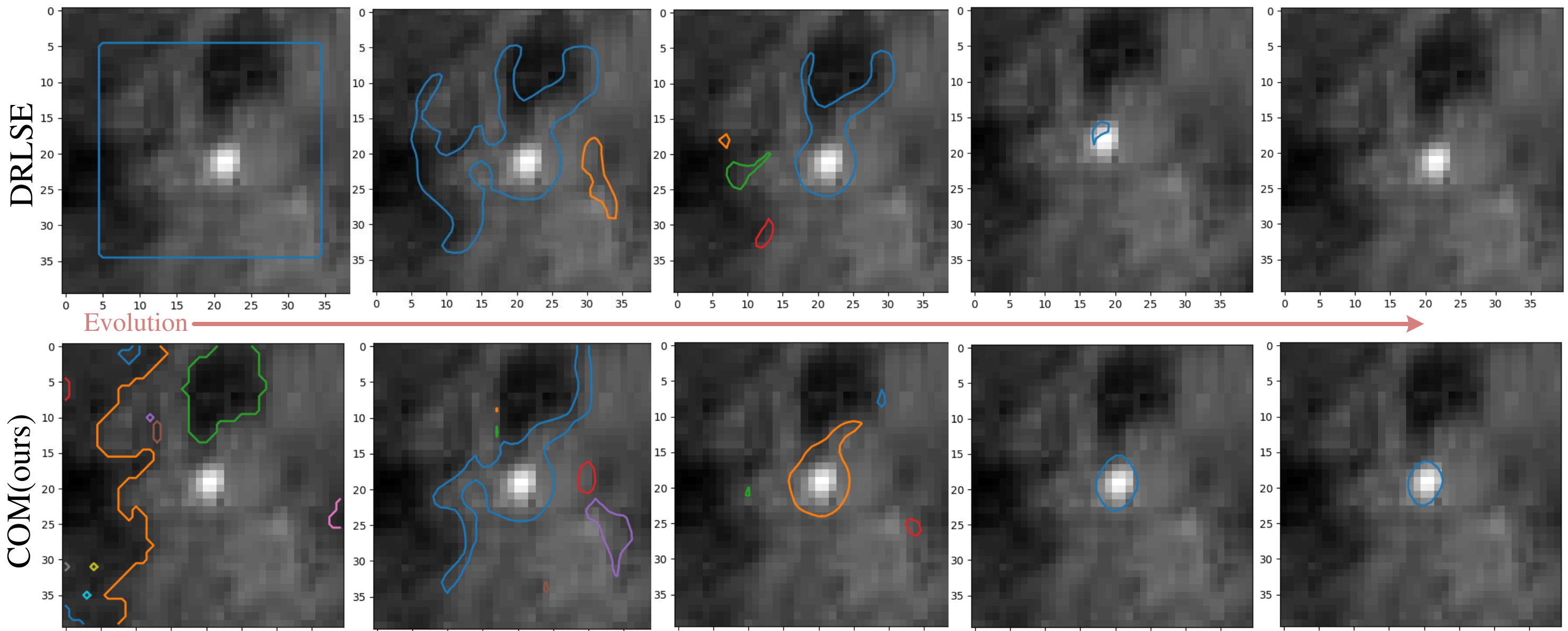}
	\caption{Qualitative contrast with the vanilla level set approach. Vanishing zero level contours exist in the vanilla DRLSE.}
	\label{Fig1}
\end{figure}

To solve these issues, several weakly supervised methods \cite{18}, \cite{19}, \cite{20}, \cite{21} have been proposed for image segmentation and detection where the masks having inexact annotation are used. Ying \textit{et al.} \cite{22} made the first attempt to achieve IRSTD with point supervision, and proved the feasibility and effectiveness. Whereas, there is a big performance gap between the weakly and fully supervised approaches. Moreover, it is extremely arduous to achieve the centroid of an infrared small target manually. Though if the point is not the centroid, it is still challenging to obtain a point annotation on a tiny infrared target exactly. It raises a question: can we propose a labor-efficient and cursory annotation approach without a considerable performance gap?

\begin{figure*}[!t]
	\centering
	\includegraphics[width=2\columnwidth]{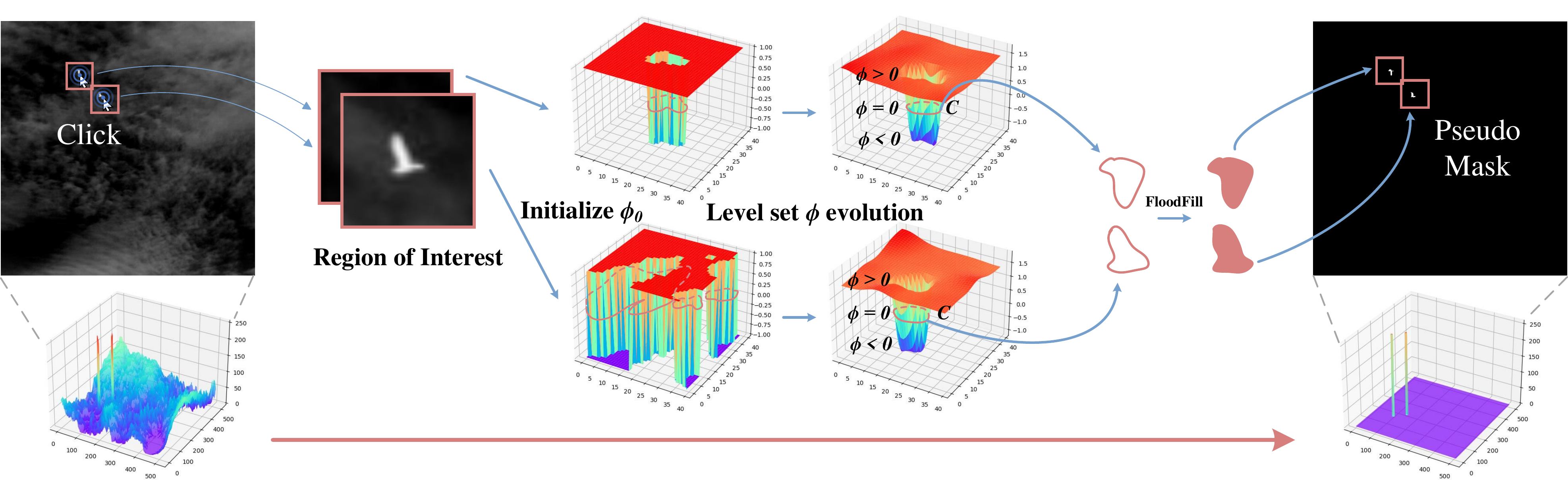}
	\caption{Overview of the proposed labor-efficient annotation framework with level set.}
	\label{Fig2}
\end{figure*}

In this letter, an original infrared image is regarded as an initial level set \cite{23}, \cite{24}, \cite{25}, \cite{26}, on the contrary, an infrared small target mask is a converged level set, as shown in Fig. \ref{Fig2}. Therefore, this letter regards the mapping from infrared image to mask as a level set evolution, and proposes an interactive annotation approach with level set, called Click on Mask (COM).

Conventional level set methods \cite{26} are not suitable for the infrared small target annotation. Due to the small target size and the strong smoothing effect of the distance regularization, it is very easy to make the zero level contour disappear, as shown in the first line of Fig. \ref{Fig1}. We observed that the infrared small targets are high-intensity areas under a clutter background, and the intensity difference with the neighborhood is particularly significant, as shown in Fig. \ref{Fig3}. Motivated by this difference, we use a binary step function with a threshold to initialize the level set, which can roughly locate the potential target region. In addition, an energy functional about this expectation difference is designed, i.e., a line integral along the zero level contour. Minimizing the energy functional is equivalent to approximating the optimal infrared small target boundary. Furthermore, to impose a greater penalty on the level set evolution when zero level contour shrinks beyond the target boundary, we proposed a signed energy coefficient based on the expectation difference.

The main contributions are summarized as:

\begin{itemize}
	\item[1)] A labor-efficient annotation framework with level set is proposed for infrared small target detection. A high-quality pseudo mask can be obtained with only one cursory click.
	
	\item[2)] A dedicated energy functional for infrared small target detection is designed. Concretely, the expectation difference energy, binary step initialization with threshold, and signed energy coefficient are designed, according to the intensity expectation difference between the areas around the target. It solves the vanishing zero level contour caused by the small size and over-regularization.
\end{itemize}

\section{Proposed Method}

\subsection{Initialization}
In the proposed approach, the two-dimensional infrared image region $I(x)$ on a domain $\Omega$ is represented as a higher-dimensional time-dependent Level Set Function (LSF) $\phi(\mathbf{x}, t)$ with spatial parameter $\mathbf{x}$ and temporal variable $t\in[0, \infty)$. The edge of the mask is the desired zero level contour $C$ , which is obtained by the LSF evolution. According to the priori, the small targets present a relatively high intensity on the infrared image, so we initialize the LSF:

\begin{equation}
	\phi_0(\mathbf{x})=\left\{
	\begin{aligned}
		-c_0, \quad \text{if}\enspace I(\mathbf{x})>i \\
		c_0, \quad \enspace\text{otherwise},
\end{aligned}
\right.
\end{equation}
where $c_0$ is a constant set to 1.0 in Section \ref{sec:Experiments}, and $i$ is set to 50.

\subsection{Energy Formulation with Expectation Difference Term}
The level set evolution is derived as the gradient flow that minimizes an energy functional with a distance regularization term \cite{26} and an external energy. However, due to the small target area and strong smoothing effect of the regularization term, the existing external energy functional tends to flatten the LSF and eventually make the zero level contour disappear, as shown in Fig. \ref{Fig1}.

\begin{figure}[!t]
	\centering
	\includegraphics[width=\columnwidth]{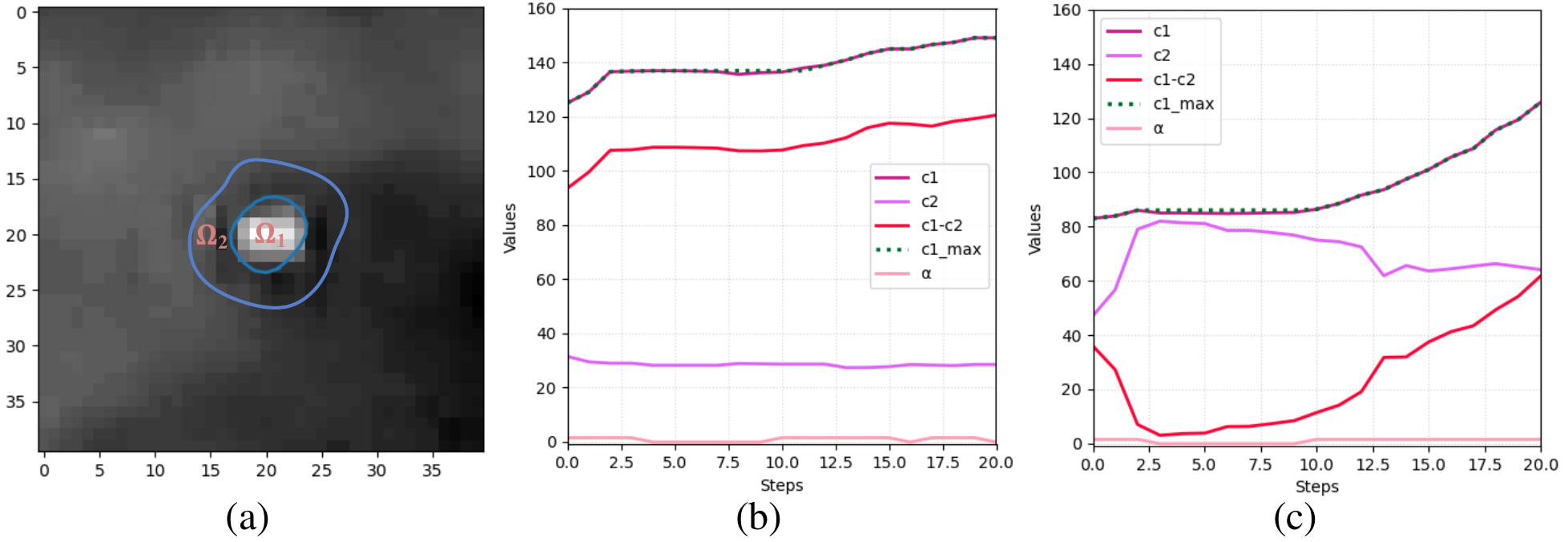}
	\caption{The intensity expectation difference between the areas around the target (sample average on the (b) IRSTD-1k and (c) NUAA-SIRST datasets).}
	\label{Fig3}
\end{figure}

To solve the issue, the Expectation Difference (ED) term is designed according to the different intensity expectations of the areas around the infrared small target (as shown in Fig. \ref{Fig3}). Our ED term is the external energy that drives the motion of the zero level set toward the desired contour. The ED between the internal and external intensity is the largest when the zero level contour $C$ is the target boundary, because the infrared small target region $\Omega_1$ has a relatively high intensity. If over-regularization exists, as shown in Fig. \ref{Fig1}, the difference decreases. Therefore, the ED gets the maximum point when the LSF evolves to the desired zero level contour $C$. We define an expectation difference indicator functional $e(\phi)$:

\begin{equation}
	e(\phi)\triangleq \frac{1}{(c_1(\phi)-c_2(\phi))^2+\Delta},
\end{equation}
where
\begin{equation}
	c_1(\phi)=\frac{\int_{\Omega_1} I(\mathbf{x})H(\phi(\mathbf{x}))d\mathbf{x}}{\int_{\Omega_1} H(\phi(\mathbf{x}))d\mathbf{x}}
\end{equation}
and
\begin{equation}
	c_2(\phi)=\frac{\int_{\Omega_2} I(\mathbf{x})H(-\phi(\mathbf{x}))d\mathbf{x}}{\int_{\Omega_2} H(-\phi(\mathbf{x}))d\mathbf{x}}
\end{equation}
are intensity expectations of the region $\Omega_1\triangleq\{\mathbf{x}|\phi(\mathbf{x})<0\}$ and adjacent region $\Omega_2$, respectively. $H(.)$ is an approximate Heaviside function \cite{27}. $\Delta$ denotes a minimal constant to prevent division by 0. The adjacent region $\Omega_2$ is defined as an area less than 3 pixels away from the zero level contour $C$ (as shown in Fig. \ref{Fig3} (a)), to avoid high-intensity background interference at a long distance.

The proposed ED term $\mathcal{E}_e(\phi)$ computes the line integral of the functional $e(\phi)$ along the zero level contour $C$:

\begin{equation}
	\mathcal{E}_e(\phi)\triangleq\int_\Omega{e(\phi)\delta(\phi)|\nabla\phi|d\mathbf{x}}, 
\end{equation}
where $\delta(.)$ is an approximate Dirac delta function \cite{27}.  $\nabla$ denotes the gradient operator. The energy functional $\mathcal{E}_e(\phi)$ is minimized when $C$ is at the precise edge of the target \cite{26}, \cite{28}.

The final energy functional $\mathcal{E}(\phi)$ can be further expressed as:

\begin{equation}
	\mathcal{E}(\phi)=\mu\mathcal{R}(\phi)+\alpha sign(c_1(\phi)-c_{1max}(\phi))\mathcal{A}(\phi)+\beta\mathcal{E}_e(\phi)
\end{equation}
where $\mu$, $\alpha$, $\beta$ are coefficients, $\mathcal{A}(\phi)=\int_\Omega\frac{H(-\phi(\mathbf{x}))}{1+|\nabla{I(\mathbf{x})}|^2}d\mathbf{x}$ and $\mathcal{R}(\phi)$  are the area energy functional and the distance regularization term defined by Li \textit{et al.} \cite{26}. $sign(.)$ is a sign function. This signed coefficient gets a positive value when $c_1(\phi)$ is greater than the current maximum value $c_{1max}(\phi)$, and the zero level contour $C$ shrinks. Otherwise, it gets a negative value (indicating that the target area tends to disappear), and $C$ expands.

\begin{figure*}[!t]
	\centering
	\includegraphics[width=2\columnwidth]{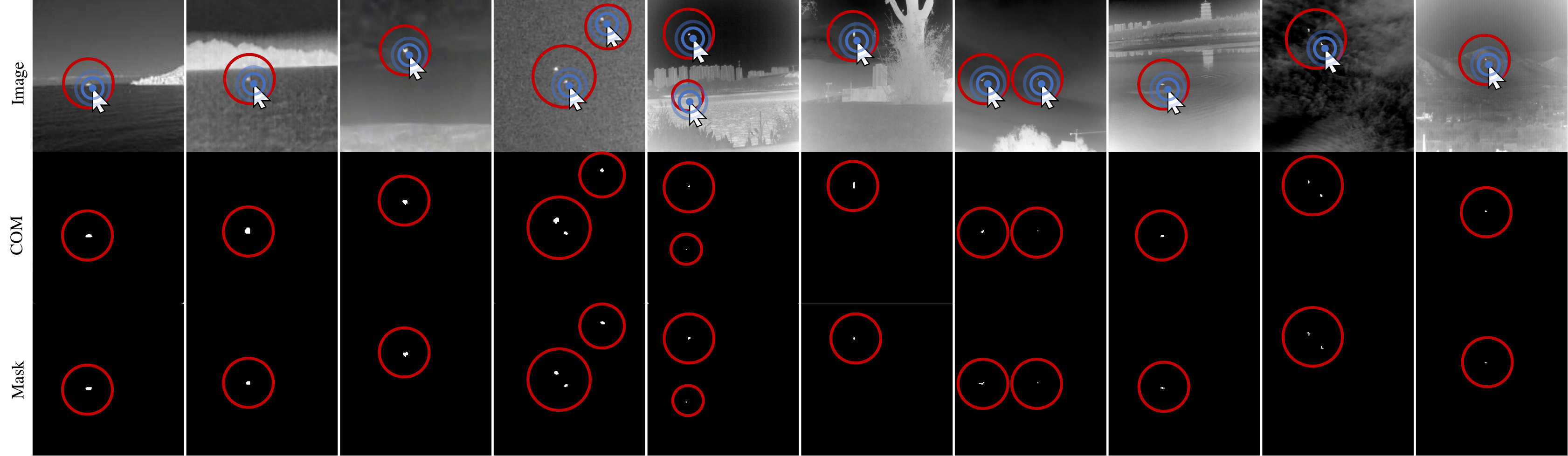}
	\caption{Visual comparison between our COM and the manual ground truth.}
	\label{Fig4}
\end{figure*}

\subsection{Gradient Flow for Energy Minimization}
Minimize the energy functional $\mathcal{E}(\phi)$ by finding the steady state solution of a gradient flow equation \cite{29}, the final zero level contour $C$ can be obtained. The gradient flow of the energy functional $\mathcal{E}(\phi)$ can be formulated as:

\begin{equation}
	\begin{aligned}
		\frac{\partial{\phi}}{\partial{t}}&=\mu div(d_p(|\nabla|\phi)|\nabla|\phi)\\
		&+\alpha sign(c_1(\phi)-c_{1max}(\phi))\int_\Omega\frac{H(-\phi(\mathbf{x}))}{1+|\nabla{I(\mathbf{x})}|^2}d\mathbf{x}\\
		&+\beta\delta(\phi)div\left(\frac{\nabla\phi}{(c_1(\phi)-c_2(\phi))^2|\nabla\phi|}\right)
	\end{aligned}
\end{equation}
where $d_p(.)$ is a variant function related to the double-well potential \cite{26} and $div(.)$ is the divergence operator.

\begin{figure}[!t]
	\centering
	\includegraphics[width=\columnwidth]{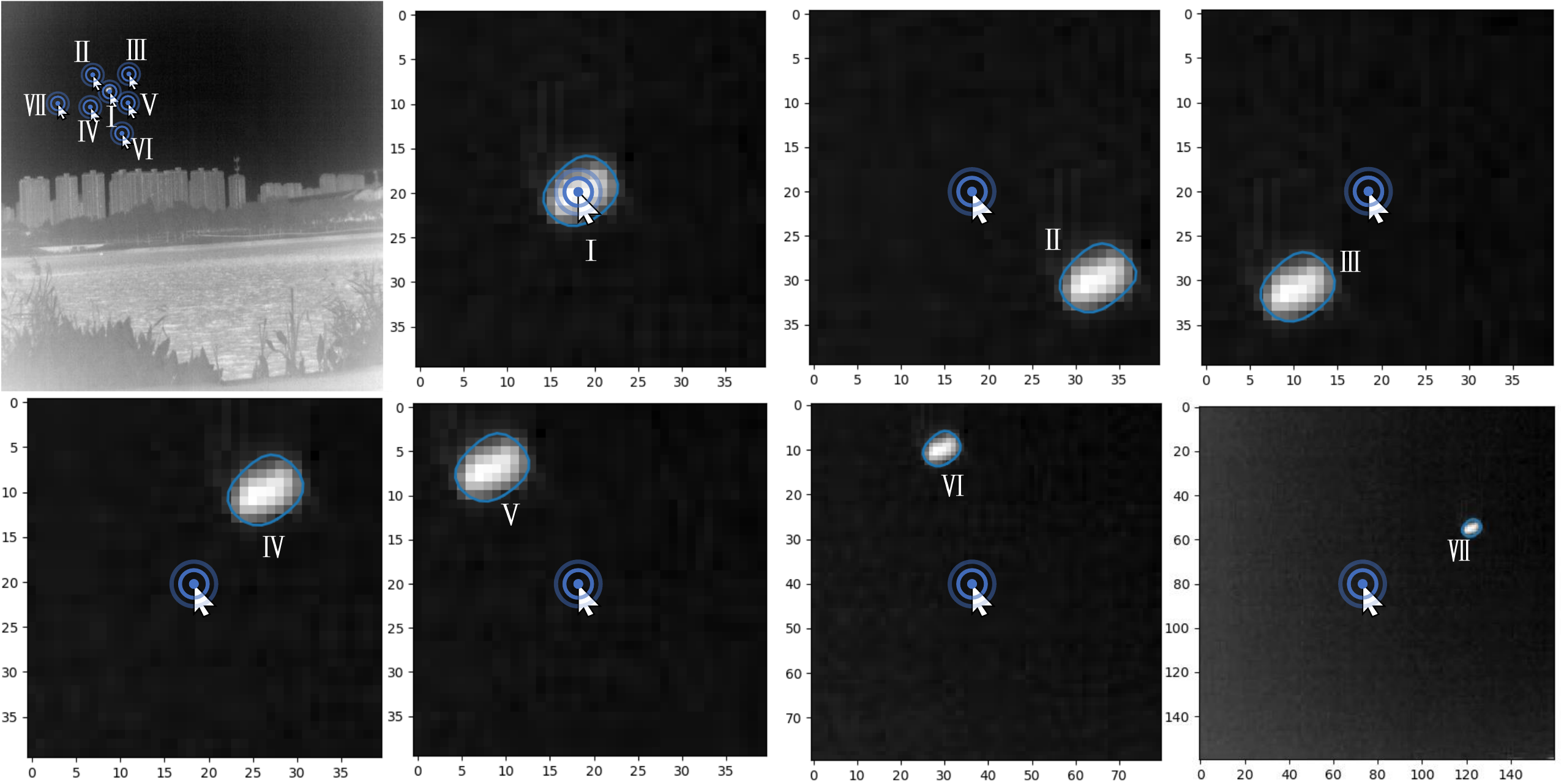}
	\caption{The arbitrariness of clicks and stability of pseudo masks. Our method allows broad-range clicks without affecting the results.}
	\label{Fig5}
\end{figure}

\begin{table}[t]
	\centering
	\caption{Comparisons between our COM annotation and manual ground truth with SOTA methods on the NUAA-SIRST and IRSTD-1K datasets (by $IoU (\%)$, $Pd (\%)$ and $Fa (10^{-6})$).}
	\label{t1}
	\resizebox{\linewidth}{!}{
		\begin{tabular}{c|ccc|ccc}
			\Xhline{1pt}
			\multirow{2}{*}{Methods}&\multicolumn{3}{c|}{NUAA-SIRST}&\multicolumn{3}{c}{IRSTD-1K}\\
			&$IoU$&$F_a\downarrow$&$P_d$&$IoU$&$F_a\downarrow$&$P_d$\\ 
			\Xhline{0.8pt}
			ACM Manual \cite{14}&63.03&8.07&95.41&50.54&2.57&84.93\\
			\rowcolor{gray!20}\textbf{ACM \textit{w} COM} &58.53&8.25&93.58&47.57&2.31&84.93\\
			\Xhline{0.8pt}
			DNA-Net Manual \cite{16}&70.04&5.01&98.17&62.11&6.81&91.78\\
			\rowcolor{gray!20}\textbf{DNA-Net \textit{w} COM} &65.97&7.23&96.33&53.53&3.89&88.36\\
			\Xhline{0.8pt}
			ISNet Manual \cite{17}&71.04&10.78&96.33&53.99&4.43&85.27\\
			\rowcolor{gray!20}\textbf{ISNet \textit{w} COM} &61.71&16.19&92.66&48.35&3.05&84.59\\
			\Xhline{0.8pt}
			AGPC Manual \cite{31}&70.91&2.62&98.17&61.32&4.29&89.04\\
			\rowcolor{gray!20}\textbf{AGPC \textit{w} COM} &65.12&2.53&97.25&51.91&3.22&85.96\\
			\Xhline{0.8pt}
			UIUNet Manual \cite{32}&71.45&8.78&97.25&62.05 &10.49&92.81\\
			\rowcolor{gray!20}\textbf{UIUNet \textit{w} COM} &61.76&12.22&93.58&53.79&3.64&85.96\\
			\Xhline{0.8pt}
			ILNet Manual \cite{30}&71.75&3.99&98.17&62.74&4.14&91.78\\
			\rowcolor{gray!20}\textbf{ILNet \textit{w} COM} &64.27&6.79&94.50&53.96&2.92&87.33\\
			\Xhline{0.8pt}
			Average Difference&-6.81&+2.33&-2.60&-7.27&-2.28&-3.08\\
			\Xhline{1pt}
		\end{tabular} 
	}
\end{table}

\begin{table}[t]
	\centering
	\caption{Comparisons with SOTA non-learning and weakly-supervised methods on the NUAA-SIRST and IRSTD-1K datasets (by $IoU (\%)$, $Pd (\%)$ and $Fa (10^{-6})$). The best results are in \textbf{BOLD}.}
	\label{t2}
	\resizebox{\linewidth}{!}{
		\begin{tabular}{c|ccc|ccc}
			\Xhline{1.2pt}
			\multirow{2}{*}{Methods}&\multicolumn{3}{c|}{NUAA-SIRST}&\multicolumn{3}{c}{IRSTD-1K}\\
			&$IoU$&$F_a\downarrow$&$P_d$&$IoU$&$F_a\downarrow$&$P_d$\\ 
			\Xhline{1pt}
			WSLCM \cite{7}&1.16&5446&77.95&3.45&6619&72.44\\
			TLLCM \cite{8}&1.03&5899&79.09&3.31& 6738& 77.39\\
			Max-Median \cite{5}&4.17&55.33&69.20&7.00&59.73&65.21\\
			Top-Hat \cite{6}&7.14&1012&79.84&10.06&1432&75.11\\
			MSLSTIPI \cite{12}&10.30&1131&82.13&11.43&1524&79.03\\
			RIPT \cite{10}&11.05&22.61&79.08&14.11&28.31& 77.55\\
			NRAM \cite{9}&12.16&13.85&74.52&15.25&16.93& 70.68\\
			PSTNN \cite{11}&22.40&29.11&77.95&24.57&35.26&71.99\\
			IPI \cite{13}&25.67&11.47&85.55&27.92&16.18&81.37\\
			\Xhline{1pt}
			ACM \textit{w} Coarse \cite{22} &47.81&40.75&88.21&40.37&64.81&\textbf{92.59}\\
			ACM \textit{w} Centroid \cite{22}&49.23&40.95&89.35&41.44&60.46&88.89\\
			ACM \textit{w} DRLSE \cite{26} &50.31&15.29&90.82&41.51&7.26&81.16\\
			\rowcolor{gray!20}\textbf{ACM \textit{w} COM (ours)}&\textbf{58.53}&\textbf{8.25}&\textbf{93.58}&\textbf{47.57}&\textbf{2.31}&84.93\\
			\Xhline{1pt}
			DNA-Net \textit{w} Coarse \cite{22} &61.13&11.87&93.16&49.05 &15.07&87.54\\
			DNA-Net \textit{w} Centroid \cite{22}&61.95&18.17&92.02&52.09&16.09&88.88\\
			DNA-Net \textit{w} DRLSE \cite{26} &53.36&16.20&89.81&44.95&10.12&83.22\\
			\rowcolor{gray!20}\textbf{DNA-Net \textit{w} COM (ours)}&\textbf{65.97}&\textbf{7.23}&\textbf{96.33}&\textbf{53.53}&\textbf{3.89}&\textbf{92.12}\\
			\Xhline{1.2pt}
		\end{tabular} 
	}
\end{table}

\section{Experiments}
\label{sec:Experiments}

We obtain the pseudo annotation on the NUAA-SIRST \cite{14} and IRSTD-1k \cite{17} training datasets, as shown in Fig. \ref{Fig4}, and apply them to the state-of-the-art methods \cite{14}, \cite{16}, \cite{17}, \cite{30}, \cite{31}, \cite{32} to demonstrate the effectiveness of the proposed annotation approach. Then we compare the COM with several non-learning methods \cite{5}, \cite{6}, \cite{7}, \cite{8}, \cite{9}, \cite{10}, \cite{11}, \cite{12}, \cite{13} and weakly-supervised methods \cite{22}. In addition, ablation studies are performed to demonstrate the effectiveness of the proposed energy functional.

\subsection{Implementation Details}
A pixel-level metric: Intersection over Union (IoU) \cite{33}; and two target-level metrics: Probability of Detection (Pd) and False-alarm rate (Fa) \cite{16} are used to quantitatively evaluate the performance.

For the approach proposed in this letter: $c_0$, $\mu$, $\beta$ are set to 1, 0.2, and 10$\Delta$, respectively. $\alpha$ is set to 1.5 or -0.1. For all the supervised methods: input images are resized to 256 × 256. All the methods are trained 300 epochs with a batch size of 8. The learning rate strategy and data augmentation are unified to make a fair comparison. Other configurations follow the original paper. All methods are implemented based on Pytorch 1.10.0 \cite{34} and conducted on an NVIDIA GeForce RTX 3080.

\subsection{Comparison with State-of-the-Art Approaches}
The COM is applied to the state-of-the-art methods \cite{14}, \cite{16} and the effectiveness is quantitatively demonstrated, as shown in Table \ref{t1}. Our approach is labor-efficient, without a significant performance gap compared with manual annotation. On the NUAA-SIRST dataset, compared with the manual annotation, the average difference of the pixel-level metric is only -6.81 IoU, while the target-level metrics are +2.33 Fa and -2.60 Pd. On the IRSTD-1k dataset, the average quantitative results are -7.27 IoU, -2.28 Fa, and -3.08 Pd. Compared with the manually annotated ground truth, our labor-efficient COM annotation actually achieves a lower false alarm rate. Experiments reveal that our pseudo masks can be effectively applied to all SOTA models, which do not have a large performance gap.

Moreover, the COM achieves great performance advantages compared with the non-learning and the weakly-supervised method, as shown in Table \ref{t2}.

\subsection{Ablation Study}
We performed ablation studies on the IRSTD-1k dataset, as shown in Table \ref{t3}. For each experimental case, we randomly sampled 48 pseudo annotations from the training dataset and calculated the average IoU with the manual ground truth. The binary step initialization with a threshold for target pre-location avoids the uncertainty about the long-distance evolution. In addition, our expectation difference term significantly enhances the performance of the level set approach. This phenomenon powerfully supports our hypothesis of the expectation difference (see section I). The effect of the signed coefficient cannot be highlighted by the IoU metric, because of the tiny areas of the targets. Vanishing zero level contour cannot make the IoU change significantly. Correspondingly, we qualitatively analyze the effect of the signed coefficient and the advantages of the level set annotation approach, to demonstrate the effectiveness of the proposed energy functional. As shown in the first line of Fig. \ref{Fig1}, the zero level contour $C$ is prone to disappear without the expectation difference term (the vanilla DRLSE). Our approach (the second line of Fig. \ref{Fig1}) effectively alleviates this issue. When the zero level contour $C$ tends to disappear, it will oscillate around the desired contour.

\subsection{Arbitrariness of Clicks and Stability of Pseudo Labels}
The interactive click during our annotation has a remarkable degree of freedom, and is not limited to the centroid points or even the targets \cite{22}. The COM allows interactive clicks outside (nearby) the target, while the final result is not affected, as shown in Fig. \ref{Fig5}. This is due to our threshold initialization of the LSF $\phi$, which locates the approximate target area at the beginning of the evolution. Such initialization reduces the evolution times and gives the interactive click sufficient arbitrariness. Unexpectedly, our COM even gets the same high-quality pseudo mask with an extremely long-range ($>$ 50 pixels) click. This incredible arbitrariness makes our COM more labor-efficient, and greatly reduces the time cost and resource consumption.

\begin{table}[!t]
	\centering
	\caption{Ablation: Contribution of our proposed Initialization, Signed Coefficient, and the ED term. Line zero indicates the vanilla DRLSE.}
	\label{t3}
	\begin{tabular}{ccccc}
		\Xhline{1.2pt}
		\#&Initialization&Signed Coefficient&ED term&$IoU(\%)$\\ 
		\Xhline{1pt}
		DRLSE& & & &43.86\\
		1&\checkmark& & &46.39\\
		2&\checkmark&\checkmark& &46.98\\
		3&\checkmark&\checkmark&\checkmark&\textbf{53.24}\\
		\Xhline{1.2pt}
	\end{tabular} 
\end{table}

\section{Conclusion}

In this letter, a labor-efficient annotation framework with level set is proposed for infrared small target detection, to reduce the annotation consumption. A high-quality pseudo mask can be obtained with only one cursory click. Exhaustive experimental comparisons over two datasets demonstrate the advantages of the proposed COM approach.


\begin{thebibliography}{34}
	
	\bibitem{1} P. Demosthenous, C. Pitris, and J. Georgiou, ``Infrared fluorescence based cancer screening capsule for the small intestine,'' {\em IEEE Trans. Biomed. Circuits Syst.}, vol. 10, no. 2, pp. 467--476, Apr. 2016. 
	
	\bibitem{2} M. Zhao, W. Li, L. Li, J. Hu, P. Ma, R. Tao, ``Single-Frame Infrared Small-Target Detection: A survey,'' {\em IEEE Geosci. Remote Sens. Mag.}, vol. 10, no. 2, pp. 87--119, Jun. 2022.
	
	\bibitem{3} J. Lu, Y. He, H. Li, and F. Lu, ``Detecting small target of ship at sea by infrared image,'' in {\em IEEE Int. Conf. Automat. Sci. Eng.}, 2006, pp. 165--169.
	
	\bibitem{4} F. Wenxing, S. Z., and K. Z., ``Conceptual research on application of single pixel infrared imaging to missile guidance,'' in {\em Asia Pacific Conf. Inform. Process.}, 2009, pp. 155--188.
	
	\bibitem{5} S. D. Deshpande, M. H. Er, R. Venkateswarlu, and P. Chan, ``Max-mean and max-median filters for detection of small targets,'' {\em Signal Data Process. Small Targets}, vol. 3809, pp. 73--84, Oct. 1999.
	
	\bibitem{6} J. Rivest and R. Fortin, ``Detection of dim targets in digital infrared	imagery by morphological image processing,'' {\em Opt. Eng.}, vol. 35, no. 7, pp. 1886--1893, Jul. 1996.
	
	\bibitem{7} J. Han, S. Moradi, I. Faramarzi, H. Zhang, Q. Zhao, X. Zhang, and N. Li, ``Infrared small target detection based on the weighted strengthened local contrast measure,'' {\em IEEE Geosci. Remote Sens. Lett.}, vol. 18, no. 9, pp. 1670--1674, Sep. 2021.
		
	\bibitem{8} J. Han, S. Moradi, I. Faramarzi, C. Liu, H. Zhang, and Q. Zhao, ``A local
	contrast method for infrared small-target detection utilizing a tri-layer window,'' {\em IEEE Geosci. Remote Sens. Lett.}, vol. 17, no. 10, pp. 1822--1826, Oct. 2019.
				
	\bibitem{9} L. Zhang, L. Peng, T. Zhang, S. Cao, and Z. Peng, ``Infrared small target detection via non-convex rank approximation minimization joint l2, 1 norm,'' {\em Remote Sens.}, vol. 10, no. 11, p. 1821, Nov. 2018.
					
	\bibitem{10} Y. Dai and W. Y., ``Reweighted infrared patch-tensor model with both nonlocal and local priors for single-frame small target detection,'' {\em IEEE J. Sel. Top. Appl. Earth Obs. Remote Sens.}, vol. 10, no. 8, pp. 3752--3767, Aug. 2017.
						
	\bibitem{11} L. Zhang and Z. Peng, ``Infrared small target detection based on partial sum of the tensor nuclear norm,'' {\em Remote Sens.}, vol. 11, no. 4, p. 382, Feb. 2019.
							
	\bibitem{12} Y. Sun, Y. J., and A. W., ``Infrared dim and small target detection via multiple subspace learning and spatial-temporal patch-tensor model,'' {\em IEEE Geosci. Remote Sens.}, vol. 59, no. 5, pp. 3737--3752, May 2020.
					
	\bibitem{13} C. Gao, D. Meng, Y. Yang, Y. Wang, X. Zhou, and A. G. Hauptmann, ``Infrared patch-image model for small target detection in a single image,'' {\em IEEE Trans. Image Process.}, vol. 22, no. 12, pp. 4996--5009, Dec. 2013.
						
	\bibitem{14} Y. Dai, Y. Wu, F. Zhou, and K. Barnard, ``Asymmetric contextual modulation for infrared small target detection,'' in {\em Proc. IEEE/CVF Winter Conf. Appl. Comput. Vis.}, 2021, pp. 950--959.
							
	\bibitem{15} Y. Dai, Y. Wu, F. Zhou, and K. Barnard, ``Attentional local contrast networks for infrared small target detection,'' {\em IEEE Trans. Geosci. Remote Sens.}, vol. 59, no. 11, pp. 9813--9824, Nov. 2021.
								
	\bibitem{16} B. Li, C. Xiao, L. Wang, Y. Wang, Z. Lin, M. Li, W. An, and Y. Guo, ``Dense nested attention network for infrared small target detection,'' {\em IEEE Trans. Image Process.}, vol. 32, pp. 1745--1758, Aug. 2023.
	
	\bibitem{17} M. Zhang, R. Zhang, Y. Yang, H. Bai, J. Zhang, and J. Guo, ``ISNet: Shape matters for infrared small target detection,'' in {\em Proc. IEEE Conf. Comput. Vis. Pattern Recognit.}, 2022, pp. 877--886.
	
	\bibitem{18} B. Cheng, O. Parkhi, and A. Kirillov, ``Pointly-supervised instance segmentation,'' in {\em Proc. IEEE Conf. Comput. Vis. Pattern Recognit.}, 2022, pp. 2617--2626.
	
	\bibitem{19} Y. Li, H. Zhao, X. Qi, Y. Chen, L. Qi, L. Wang, Z. Li, J. Sun, and J. Jia, ``Fully convolutional networks for panoptic segmentation with point-based supervision,'' {\em IEEE Trans. Pattern Anal. Mach. Intell.}, vol. 45, no. 4, pp. 4552--4568, Apr. 2022.
	
	\bibitem{20} S. Zhang, Z. Yu, L. Liu, X. Wang, A. Zhou, and K. Chen, ``Group r-cnn for weakly semi-supervised object detection with points,'' in {\em Proc. IEEE Conf. Comput. Vis. Pattern Recognit.}, 2022, pp. 9417--9426.
	
	\bibitem{21} L. Chen, T. Yang, X. Zhang, W. Zhang, and J. Sun, ``Points as queries: Weakly semi-supervised object detection by points,'' in {\em Proc. IEEE Conf. Comput. Vis. Pattern Recognit.}, 2021, pp. 8823--8832.
	
	\bibitem{22} X. Ying, L. Liu, Y. Wang, R. Li, N. Chen, Z. Lin, W. Sheng, and S. Zhou, ``Mapping degeneration meets label evolution: Learning infrared small target detection with single point supervision,'' in {\em Proc. IEEE Conf. Comput. Vis. Pattern Recognit.}, 2023, pp. 15528--15538.
	
	\bibitem{23} S. Osher and J. Sethian, ``Fronts propagating with curvature-dependent speed: Algorithms based on Hamilton-Jacobi formulations,'' {\em J. Comput. Phys.}, vol. 79, no. 1, pp. 12--49, Nov. 1988.
	
	\bibitem{24} V. Caselles, F. Catte, T. Coll, and F. Dibos, ``A geometric model for active contours in image processing,'' {\em Numer. Math.}, vol. 66, no. 1, pp. 1--31, Dec. 1993.
		
	\bibitem{25} R. Malladi, J. A. Sethian, and B. C. Vemuri, ``Shape modeling with front propagation: A level set approach,'' {\em IEEE Trans. Pattern Anal. Mach. Intell.}, vol. 17, no. 2, pp. 158--175, Feb. 1995.
	
	\bibitem{26} C. Li, C. Xu, C. Gui, and M. D. Fox, ``Distance regularized level set evolution and its application to image segmentation,'' {\em IEEE Trans. Image Process.}, vol. 19, no. 12, pp. 3243--3254, Dec. 2010.

	\bibitem{27} H. Zhao, T. Chan, B. Merriman, and S. Osher, ``A variational level set approach to multiphase motion,'' {\em J. Comput. Phys.}, vol. 127, no. 1, pp. 179--195, Aug. 1996.
	
	\bibitem{28} V. Caselles, R. Kimmel, and G. Sapiro, ``Geodesic active contours,'' {\em Int. J. Comput. Vis.}, vol. 22, no. 1, pp. 61--79, Feb. 1997.
	
	\bibitem{29} G. Aubert and P. Kornprobst, {\it Mathematical Problems in Image Processing: Partial Differential Equations and the Calculus of Variations}. New York: Springer-Verlag, 2002.
	
	\bibitem{30} H. Li, J. Yang, R. Wang, and Y. Xu , ``ILNet: Low-level matters for salient infrared small target detection,'' 2023, {\em arXiv:2309.13646}.
	
	\bibitem{31} T. Zhang, L. Li, S. Cao, T. Pu, and Z. Peng, ``Attention-guided pyramid context networks for detecting infrared small target under complex background,'' {\em IEEE Trans. Aerosp. Electron. Syst.}, vol. 59, no. 4, pp. 4250--4261, Aug. 2023.
		
	\bibitem{32} X. Wu, D. Hong, and J. Chanussot, ``UIU-Net: U-Net in U-Net for infrared small object detection,'' {\em IEEE Trans. Image Process.}, vol. 32, pp. 364--376, 2023.

	\bibitem{33} J. Yu, Y. Jiang, Z. Wang, Z. Cao, and T. Huang, ``Unitbox: An advanced object detection network,'' in {\em ACM Int. Conf. Multimedia}, 2016, pp. 516--520.

	\bibitem{34} A. Paszke, S. Gross, F. Massa, A. Lerer, J. Bradbury, G. Chanan, T. Killeen, Z. Lin, N. Gimelshein, and L. Antiga, ``Pytorch: An imperative style, high-performance deep learning library,'' in {\em Adv. Neural Inf. Process. Syst.}, 2019, 8026--8037.

\end{thebibliography}
\end{document}